\documentclass[letterpaper, 10 pt, conference]{ieeeconf}  

\IEEEoverridecommandlockouts                              

\usepackage{amsmath} 
\usepackage{amssymb}  
\usepackage{graphicx}
\usepackage{bbm}
\usepackage{amsmath}
\usepackage{algorithmic}
\usepackage{algorithm}
\usepackage{multirow} 
\usepackage{makecell}
\usepackage[flushleft]{threeparttable}
\usepackage{color}

\usepackage{caption}
\usepackage{subcaption}
\usepackage{hyperref}

\usepackage{algorithm}
\usepackage{graphicx} 

\title{\LARGE \bf
SafeAug: Safety-Critical Driving Data Augmentation from Naturalistic Datasets
}

\author{Zhaobin Mo$^{1,*}$, Yunlong Li$^{2,*}$, and Xuan Di$^{1,3, \dag}$
\thanks{This work is sponsored by NSF CPS-2038984 and NSF ERC-2133516.}
\thanks{$^{*}$Equal contributions; $^{\dag}$Corresponding author;}
\thanks{$^{1}$Zhaobin Mo and Xuan Di are affiliated with the Department of Civil Engineering and Engineering Mechanics, Columbia University, 500 West 120th Street, New York, NY 10025, USA
        {\tt\small zm2302@columbia.edu, sharon.di@columbia.edu}}
\thanks{$^{2}$Yunlong Li is affiliated with the Department of Electrical Engineering, Columbia University, 500 West 120th Street, New York, NY 10025, USA
        {\tt\small yl5330@columbia.edu}}
\thanks{$^{3}$Xuan Di is also affiliated with the Data Science Institute, Columbia University, 550 W 120th St, New York, NY 10027, USA}}

\begin{document}

\maketitle
\thispagestyle{empty}
\pagestyle{empty}

\begin{abstract}
Safety-critical driving data is crucial for developing safe and trustworthy self-driving algorithms. Due to the scarcity of safety-critical data in naturalistic datasets, current approaches primarily utilize simulated or artificially generated images. However, there remains a gap in authenticity between these generated images and naturalistic ones. We propose a novel framework to augment the safety-critical driving data from the naturalistic dataset to address this issue. In this framework, we first detect vehicles using YOLOv5, followed by depth estimation and 3D transformation to simulate vehicle proximity and critical driving scenarios better. This allows for targeted modification of vehicle dynamics data to reflect potentially hazardous situations. Compared to the simulated or artificially generated data, our augmentation methods can generate safety-critical driving data with minimal compromise on image authenticity. Experiments using KITTI datasets demonstrate that a downstream self-driving algorithm trained on this augmented dataset performs superiorly compared to the baselines, which include SMOGN and importance sampling.


\end{abstract}

\section{Introduction}

As the adoption of autonomous driving technology accelerates, it is critical to ensure the safety and reliability of self-driving vehicles (AV) \cite{su2023uncertainty}. Real-world driving environments are highly dynamic and unpredictable, so self-driving cars must effectively handle safety-critical data-anomalous or extreme situations not commonly found in safety-critical data. 


Current research in training algorithms for autonomous driving focuses on three main techniques—simulation, image generation, and sampling-based methods—each aiming to enrich training datasets with realistic, safety-critical data. Simulation-based methods create controlled virtual environments but often struggle to capture the complex dynamics of the real world \cite{wang2023advsim, gao2023autonomous,gao2024decentralized}. Sampling-based methods, such as SMOTE \cite{Chawla_2002} and Importance sampling \cite{kloek1978bayesian}, augment dataset diversity but can miss natural driving subtleties and introduce non-authentic artifacts \cite{Ling2010}. Image generation techniques use graphics and deep learning to produce new images \cite{zhang2021provably, zhang2022extra, Zhang_2022_CVPR, Yang_2020_CVPR}, but achieving high realism and accurate real-world representation remains challenging. These limitations highlight the need for more naturalistic data augmentation methods to adapt autonomous driving systems to real-world conditions better.

This paper aims to enhance the training of autonomous driving models by introducing an image enhancement technique for augmenting driving data. Our approach modifies natural images to include realistic and challenging driving scenarios, thereby maintaining the authenticity of real-world models. We adjust the distances between vehicles in critical scenes by detecting vehicles within a driving dataset and employing depth estimation followed by 3D model reconstruction. We define events involving the highest deceleration levels as safety-critical data and use these scenarios to augment our training dataset further. This enhancement significantly improves the model's performance in critical situations, thereby increasing the safety and effectiveness of autonomous vehicle (AV) systems in managing unpredictable real-world driving conditions. The efficacy of our proposed method is evaluated using the KITTI dataset.

The rest of the paper is organized as follows: Sec. II reviews literature related to dataset augmentation; Sec. III introduces preliminaries and problem statements; Sec. IV details the methodology, including modules for depth estimation, vehicle detection, and data augmentation; Sec. V presents the experimental setup and results; Sec. VI concludes the study.

\section{Literature Review}

This section reviews existing literature related to the processing of safety-critical data in driving datasets, methods used in machine learning to address data imbalances, and data enhancement and simulation specializations aimed at augmenting driving datasets.

\subsection{Safety-critical driving data discovery}

Some technologies focus on generating synthetic data that simulates safety-critical scenarios \cite{wang2023advsim,hanselmann2022king,sarva2023adv3d,ke2024real,ouyang2022safety}, often involving potential collisions or dangerous driving conditions. These approaches use simulation to enhance the realism and applicability of scenarios to realistic situations. For example, AdvSim \cite{wang2023advsim} updates LiDAR sensor data in a simulated environment to create hazardous driving scenarios, which are critical for testing and improving vehicle response under duress. Similarly, KING \cite{hanselmann2022king} utilizes kinematic gradients in the simulation environment to generate scenarios specifically geared towards collision avoidance, thereby enhancing the model's predictive capabilities in high-risk situations. Some techniques \cite{WANG2020257} \cite{alhaija2017augmented} focus on augmenting real-world images or generating entirely new images containing challenging driving conditions. These methods change the background scene, lighting, and weather conditions and introduce occlusions to create diverse and challenging datasets.

SIMBAR \cite{Zhang_2022_CVPR} and SurfelGAN \cite{Yang_2020_CVPR} are examples of methods that manipulate sensor inputs and environmental conditions (e.g., lighting) to simulate different times of the day or weather conditions to create richer, more diverse training data. These methods aim to prepare driving models for the variability and unpredictability of real-world driving. Some studies, such as RADAR \cite{9896871} and TauAud \cite{9742083}, focus on augmenting the data by altering the driving style or testing the robustness of the image recognition system in adverse weather conditions. These approaches enrich the scenarios of the dataset, test the limits of current autonomous driving systems, and ensure that the system can effectively handle unexpected situations.

While specialized simulation techniques used for automated driver training are very beneficial, they have distinct limitations, primarily the inability to simulate real-world conditions perfectly. This discrepancy can result in models that perform well in controlled simulated environments but poorly in unexpected or uncommon real-world scenarios. The challenge is to create sufficiently varied and realistic simulations to prepare systems for the unpredictability of real-world driving and to ensure that they remain robust and effective under all possible conditions.

\subsection{Data augmentation for driving images}

Several strategies have been devised to mitigate the problems posed by unbalanced datasets, focusing mainly on enhancing the representation of a few categories during the training process of machine learning models. Data-level techniques are widely used, including resampling \cite{Chawla_2002} or data augmentation \cite{wu2023fineehr,ye2023medlens}. Resampling can increase its instances by oversampling minorities or reduce its instances by undersampling majorities. The Synthetic Minority Oversampling Technique (SMOTE) \cite{Chawla_2002} is a notable method in oversampling that creates synthetic samples rather than replicating existing samples. 

Data augmentation in image classification commonly involves transformations such as rotations, flips, and translations to create new data points. These manipulations help enhance minority categories by providing more diverse examples for model training, potentially improving the robustness of the classification model.

In the context of driving data, algorithmic-level techniques such as importance sampling \cite{kloek1978bayesian} and Cost-sensitive learning \cite{Ling2010} adjust the learning process to counteract bias against most categories. Ensemble methods like boosting and bagging also play a crucial role. These methods typically involve creating multiple subsets of data and ensuring that each subset provides a balanced view so the combined model has better generalization across different class distributions. Techniques like importance sampling specifically focus on sampling techniques that weigh the minority class more heavily during training to improve model performance in critical applications such as autonomous driving.

\section{Problem Statement}

The objective of enhancing the realism and applicability of datasets for training autonomous driving systems arises from traditional datasets often lacking sufficient representation of hazardous scenarios, which are essential for developing robust predictive models. The proposed approach augments images from the driving dataset, denoted by \( I \), to include scenarios that simulate more dangerous conditions, producing \( I_{\text{aug}} \). 
\begin{equation}
I_{\text{aug}}(i) \leftarrow I(i)
\end{equation}

The augmentation process initiates with each raw image \( I(i) \) undergoing depth estimation and vehicle detection, thereby creating depth and tagged images, respectively. These are utilized to generate a 3D model of the scene, which is then manipulated to simulate hazardous driving conditions by adjusting vehicle distances. The modified 3D model is subsequently converted back into a 2D augmented image, \( I_{\text{aug}}(i) \).

Evaluation of model performance is conducted under different scenarios:
1. Training with the original images, \( I \).
2. Training with the original images, \( I \), augmented by other methods, serving as baselines.
3. Training with a combination of original and augmented images, \( I + I_{\text{aug}} \).

Accuracy in predicting vehicle acceleration is assessed, explicitly focusing on handling safety-critical driving data, representing the most challenging driving scenarios. The analysis determines whether the inclusion of \( I_{\text{aug}} \) enhances the ability to predict critical outcomes in real-world hazardous conditions more effectively than baseline methods. The training optimization for the downstream car-following model is defined as:
\begin{equation}
\theta^* = \arg\min_{\theta} D(\theta; I + I_{\text{aug}})
\end{equation}
where \( \theta^* \) denotes the optimal model parameters, and \( D(\theta; I + I_{\text{aug}}) \) quantifies the deviation of model predictions from the true vehicle accelerations under both baseline and augmented conditions.

\section{Methodology}

\begin{figure*}[htbp]
\centering
    \includegraphics[width=0.8\linewidth]{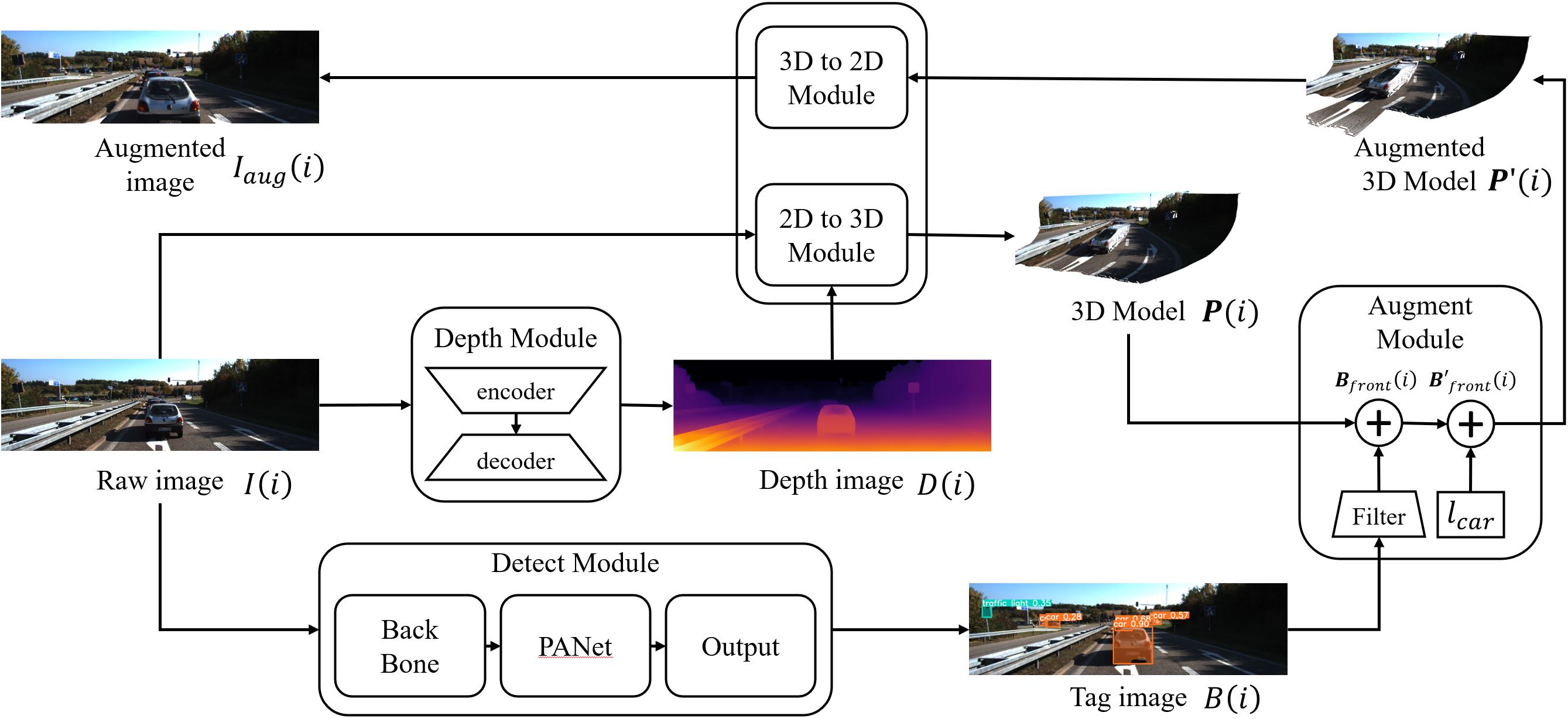}
    \caption{Autonomous Driving Data Augmentation Pipeline}
    \label{fig:pipeline}
\end{figure*}

Our research aims to enhance the training datasets for autonomous driving systems to prepare them for critical driving scenarios better. We start with the KITTI dataset, processing each image through several transformational stages as depicted in the diagram, refer to Figure \ref{fig:pipeline}.

The process involves several key transformations. First, the Detect Module is performed using the YOLOv5 framework, where the vehicle detection within each image $i \in I$ is executed. The detection process is mathematically expressed as follows:
\begin{align}
    B(i) & = \text{NMS}\left(\text{Decode}(s, a)\right) \text{where} s \in S(F(i)), a \in A
\end{align}

Where $F(i)$ is the feature map extracted from $i$, $A$ denotes the set of anchor boxes, $S$ applies convolutional operations on $F(i)$, $\text{Decode}$ computes the bounding boxes from scores and anchors, and $\text{NMS}$ is the Non-Maximum Suppression algorithm used to finalize the detection boxes.

Subsequently, Depth Estimation is carried out using an encoder-decoder model:
\begin{equation}
    D(i) = g_{\psi}(f_{\theta}(i))
\end{equation}
where $f_{\theta}$ is the encoder with parameters $\theta$, and $g_{\psi}$ is the decoder with parameters $\psi$, producing the depth map $D(i)$.

The depth maps $D(i)$ are then transformed into 3D models via the following conversion using the Open3D library:
\begin{align}
\mathbf{P}(i) = \left( \frac{(u - c_x)}{f_x} Z(u, v), \frac{(v - c_y)}{f_y} Z(u, v), Z(u, v) \right)
\end{align}
where $Z(u, v)$ is the depth value at pixel coordinates $u, v$, $c_x, c_y$ are the coordinates of the image center, and $f_x, f_y$ are the camera's intrinsic focal lengths along the x and y axes, respectively. $\mathbf{P}(i)$ denotes the 3D model created from image $i$.

In the Dataset Augmentation for Hazard Simulation step, the 3D point cloud model $\mathbf{P}(i)$ and the bounding box $\mathbf{B}_{front}(i)$ are used to adjust the position of detected vehicles to simulate hazardous scenarios:
\begin{equation}
    \mathbf{B}'_{front}(i) = \mathbf{B}_{front}(i) - \text{half body length}
\end{equation}
This adjustment simulates a reduced distance to the vehicle in front. The original acceleration data $A_{orig}$ is adjusted to $A_{aug}$, reflecting an increased risk due to closer proximity. The adjusted 3D model is then converted back into a 2D image to simulate a new, more hazardous driving scenario:
\begin{equation}
    I_{aug}(i) = \text{3D to 2D Module}(\mathbf{P}'(i))
\end{equation}
Finally, the augmented image $I_{aug}(i)$ and adjusted acceleration data $A_{aug}$ are saved in the dataset.

In parallel with these visual modifications, adjustments to the dataset's metadata, such as vehicle acceleration, align with the new risk levels depicted in the augmented images. This ensures consistency between the visual data and its associated parameters. The entire pipeline facilitates a systematic approach to generating a dataset that better prepares autonomous driving systems to handle critical driving scenarios, enhancing both the safety and robustness of these systems in real-world conditions.

\subsection{Depth Estimation Module}

\begin{algorithm}
\caption{Depth Map Generation using Depth-Anything}
\begin{algorithmic}[1]
\REQUIRE $i$: Input image from KITTI dataset
\ENSURE $D(i)$: Depth map of image $i$

\STATE Load the model from Depth-Anything
\STATE $D(i) \leftarrow \text{Depth-Anything}(i)$ 

\RETURN $D(i)$
\end{algorithmic}
\end{algorithm}

\begin{figure}[h]
    \centering
    \includegraphics[width=1\linewidth]{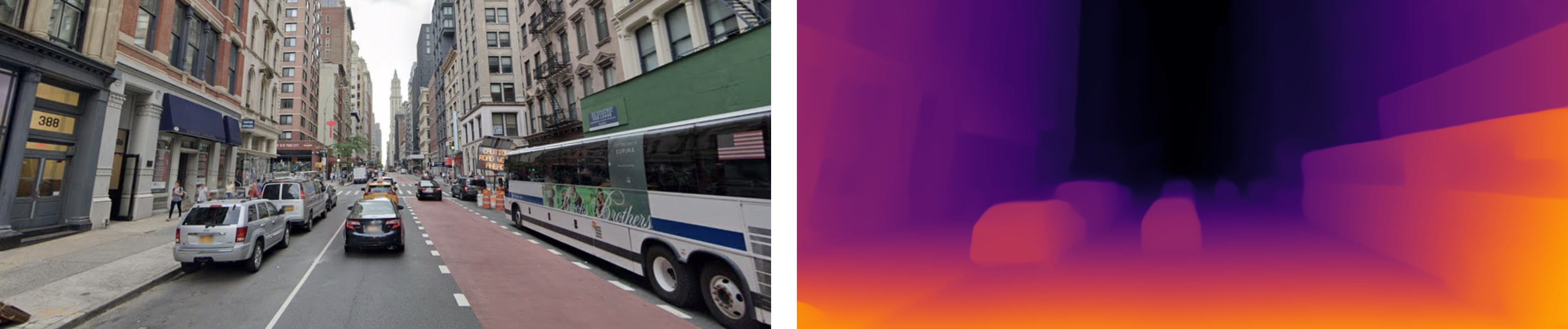}
    \caption{Street View to Depth}
    \label{fig:depth}
\end{figure}

For the depth estimation part of the project, we used Depth-Anything \cite{depthanything}, a tool known for its high accuracy and ease of use in generating depth maps from a single image. This tool was chosen because of its proven effectiveness and compatibility with our dataset, allowing us to obtain accurate depth information that is essential for realistic 3D modeling. Figure \ref{fig:depth} shows an example of a depth map generated using this tool.

\subsection{Vehicle Detect Module}
\begin{figure}[h]
    \centering
    \includegraphics[width=1\linewidth]{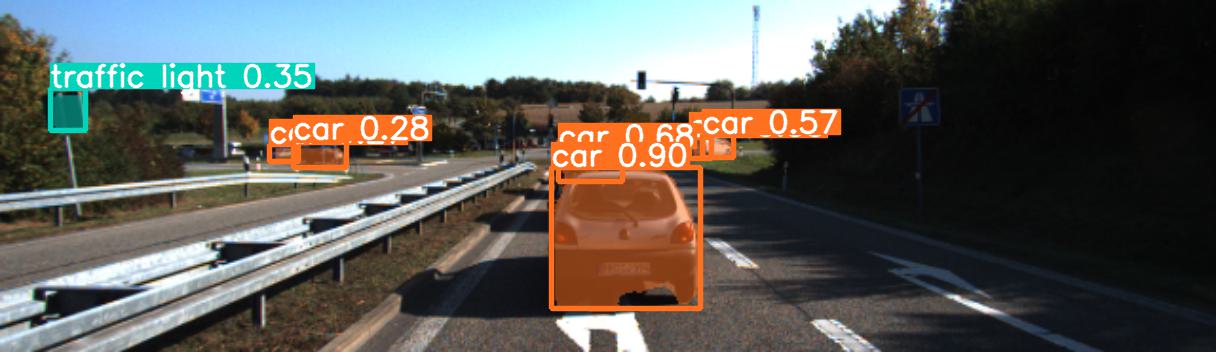}
    \caption{YOLO-v5 Result}
    \label{fig:yolov5}
\end{figure}

In the vehicle detection phase of the study, we used YOLOv5\cite{yolov5}, an object detection model known for its efficiency and accuracy. YOLOv5 was chosen for its robustness and ability to handle complex scenarios typical of autonomous driving datasets such as KITTI. An example of vehicle detection using YOLOv5 is displayed in Figure \ref{fig:yolov5}.

To determine which vehicle is located directly in front of the camera-equipped vehicle, we used an approach based on the bounding box provided by YOLOv5. We drew a virtually vertical line in the center of each image and identified the largest bounding box intersecting that line as the vehicle directly in front of it. Vehicles directly in front are identified in order to facilitate the creation of dangerous situations

\begin{algorithm}
\caption{Vehicle Detection using YOLOv5}
\begin{algorithmic}[1]
\REQUIRE $i$: Input image from KITTI dataset
\ENSURE $B_{front}$: Bounding box of the vehicle directly in front

\STATE Load the pre-trained YOLOv5 model
\STATE $B \leftarrow \text{YOLOv5}(i)$ 
\STATE $C \leftarrow \text{width}(i) / 2$
\STATE $B_{front} \leftarrow \text{null}$

\FOR{each $b$ in $B$}
    \IF{($\text{center}(b) \text{ is close to } C) \&
    (\text{not } B_{front} \text{ or } \text{area}(b)>\text{area}(B_{front}))$}
        \STATE $B_{front} \leftarrow b$ 
    \ENDIF
\ENDFOR

\RETURN $B_{front}$
\end{algorithmic}
\end{algorithm}

\subsection{3D 2D Conversion Module}

After successfully acquiring the depth maps, we used the Open3D\cite{Zhou2018} library to merge these depth maps with the corresponding raw images to create the 3D model. open3D provides a powerful set of point cloud data processing and visualization tools that help to accurately reconstruct 3D environments. Integrating the depth maps and raw images into the 3D model is critical for simulating and analyzing real-world driving scenarios, which allows us to realistically adjust the vehicle's position in the model to reflect closer, more dangerous driving conditions.

Once the 3D model has been augmented to represent more dangerous situations, the next step is to convert these modified 3D models back into 2D images. This conversion is critical for updating the dataset with new images reflecting the altered driving conditions, thus enhancing the training material for the autopilot system.

\begin{algorithm}
\caption{3D Model Creation and 2D Conversion using Open3D}
\begin{algorithmic}[1]
\REQUIRE $i$: Input raw image from KITTI dataset
\REQUIRE $D$: Corresponding depth map
\REQUIRE $B_{front}$: Bounding box of the vehicle directly in front
\ENSURE $I_{aug}$: Augmented 2D image

\STATE Load $i$ and $D$
\STATE $P \leftarrow \text{Open3D}.create\_point\_cloud(D, i)$ 
\STATE Adjust $P$ using $B_{front}$ to simulate closer vehicle 
\STATE $i_{aug} \leftarrow \text{Open3D}.project\_point\_cloud\_to\_image(P)$ 

\RETURN $i_{aug}$
\end{algorithmic}
\end{algorithm}

\subsection{Augment Module}

\begin{figure}[h]
    \centering
    \includegraphics[width=0.9\linewidth]{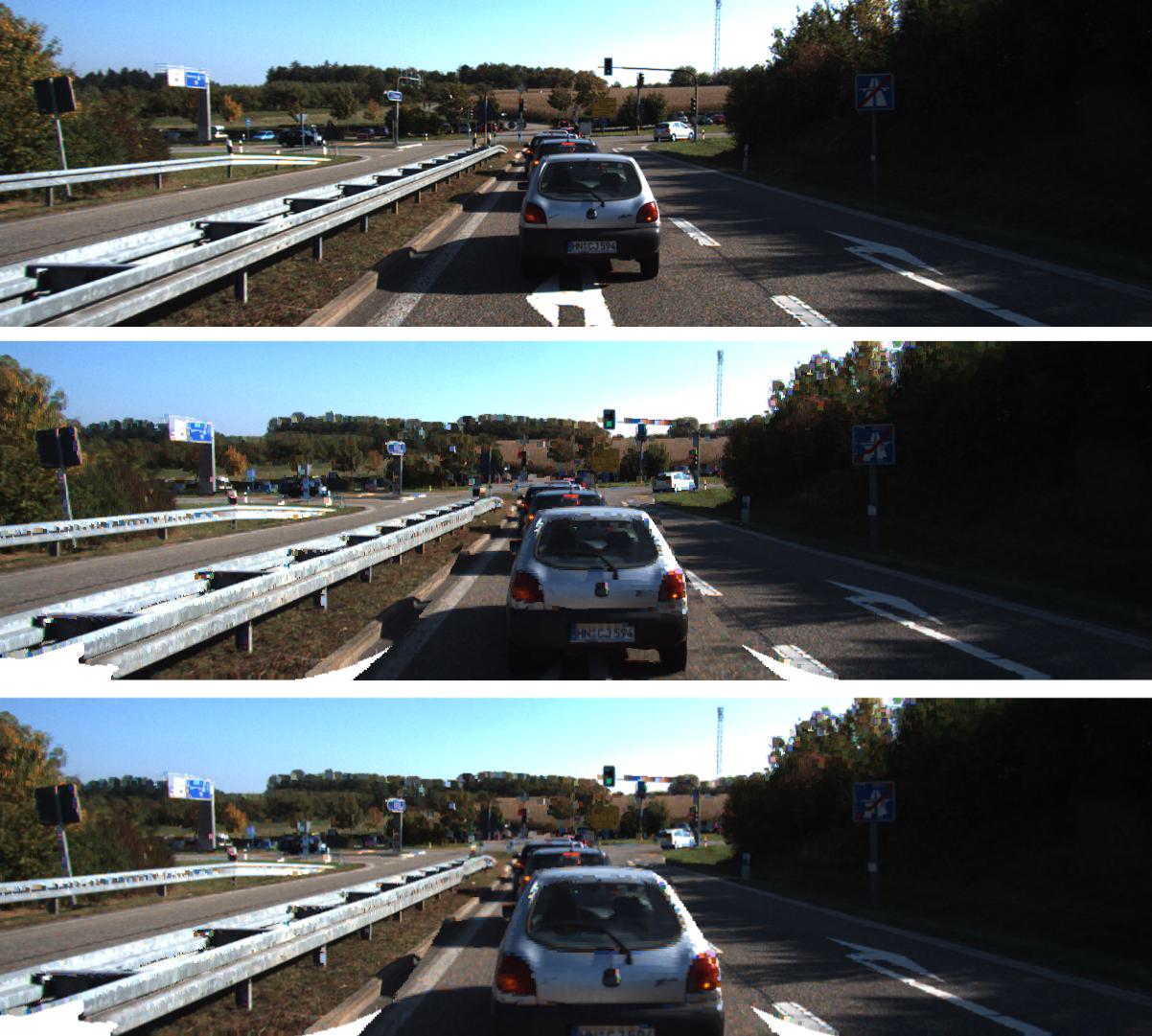}
    \caption{Comparison of Vehicle Distance Adjustments}
    \label{fig:3de22d}
\end{figure}

In the augmentation phase, we used the vehicle positions detected by YOLOv5 in the original image to adjust the distance between vehicles in the 3D model. Specifically, we moved the detected vehicle (the one directly in front of us) by half a body distance along the horizontal axis relative to the camera in 3D space. This adjustment is noticeable and significant because we selected images where the initial distance to the front vehicle was already close. The spatial adjustment is then accurately reflected in the 2D image through a 3D to 2D conversion process. As shown in Figure \ref{fig:3de22d}, we demonstrate the effects of varying the distance by different amounts, including the half-body distance change, to show its impact.

This modification not only made the vehicle appear closer in the image, simulating a more dangerous driving scenario, but it also addressed potential concerns regarding the changes in shadows and reflections. Due to the camera's perspective, any misaligned shadows are covered, maintaining the image's realism. Additionally, we did not employ any generative methods, ensuring that the 3D spatial relationships within the image were preserved.

In addition, to ensure that the augmented data reflect realistic driving conditions, we adjusted the vehicle acceleration data accordingly to represent the severe braking typically required in emergency situations. Specifically, we set the acceleration adjustment value to 1.5 times the original acceleration of each image, as the closer vehicle distance at the same speed necessitates a greater deceleration.

The above augmentation was performed on approximately 200 images and integrated back into the original dataset to create an augmented dataset. The original dataset consists of approximately 2000 images. Our goal is to evaluate whether this enhanced dataset can indeed improve the model's performance in processing and responding to high-risk driving scenarios through subsequent testing.

\begin{algorithm}
\caption{Dataset Augmentation for Hazard Simulation}
\begin{algorithmic}[1]
\REQUIRE $P$: 3D point cloud model
\REQUIRE $B_{front}$: Bounding box of the vehicle directly in front
\REQUIRE $A_{orig}$: Original acceleration data
\ENSURE $I_{aug}$: Augmented 2D image
\ENSURE $A_{aug}$: Adjusted acceleration data

\STATE $distance\_adjustment \leftarrow \text{half body length of the car}$
\STATE Move $B_{front}$ in $P$ by $distance\_adjustment$ closer to the camera
\STATE $A_{aug} \leftarrow 1.5 \times A_{orig}$ 
\STATE $I_{aug} \leftarrow \text{3D to 2D Module}(P)$ 

\STATE Save $I_{aug}$ and $A_{aug}$ in dataset 

\RETURN $I_{aug}, A_{aug}$
\end{algorithmic}
\end{algorithm}

\section{Experiment}

\subsection{Dataset}
In this study, we chose the KITTI dataset\cite{Geiger2013IJRR} because it comprehensively captures real-world driving conditions and is widely recognized in the field of autonomous driving research. Specifically, we used the ``processed (synced+rectified) color sequences" subset, which provides images at a resolution of 0.5 Megapixels in PNG format. This subset has high-quality image data and accurately synchronized images, making it particularly suitable for this study. In addition to the visual data, we also utilized the kinematic data in the dataset, specifically the acceleration and velocity of the vehicle. This information is crucial to our research as it allows us to train the model on the dynamics of the driving scenario.

\subsection{Experiment Setting}

In our study, the experimental framework revolves around a basic convolutional neural network (CNN) model. The model is designed to predict vehicle acceleration using two main inputs: images from the dataset and vehicle speed data. The choice of a relatively CNN model was intentional, as it allows for easier training and a clearer explanation of how the quality of the dataset affects model performance.

\subsubsection{Baseline Methods}

In our study, we evaluate the effectiveness of dataset augmentation in automated driving applications and compare it to two well-established techniques designed to deal with unbalanced datasets: Synthetic Minority Over-Sampling Technique for Regression with Gaussian Noise (SMOGN)  \cite{smogn} and importance sampling \cite{kloek1978bayesian}. SMOGN is a modification of the SMOTE algorithm used for regression by synthesizing new examples with additional Gaussian noise to enhance the dataset, thereby increasing the representativeness and variability of dangerous driving scenarios. This helps to balance the dataset and improve the robustness of the model. Importance sampling focuses on probabilistically adjusting the samples to over-represent rare but critical events such as emergency braking, thus ensuring that the model can effectively handle these critical scenarios. By using these methods as benchmarks, we aim to demonstrate that models trained using our augmented dataset have higher accuracy and safety.

\subsubsection{Dataset Splitting}

In preparation for training the CNN model, the dataset was split to focus on hazardous scenarios, which are critical for testing the effectiveness of an autonomous driving system in hazard-prone situations. The dataset was sorted according to the vehicle acceleration values and the 10\% of data with the smallest acceleration values and the car in front, indicating strong deceleration and potentially hazardous situations, were categorized as safety-critical data. This subset is important because it represents less frequent but more challenging and dangerous situations, thus providing a robust test of the model's ability to handle critical situations. By comparing the performance in these safety-critical data before and after the dataset expansion, we aim to verify whether adding processed images to simulate more dangerous driving conditions enhances the model's ability to handle such situations effectively.

\subsection{Results}

The performance evaluation of the CNN models is centered around two main scenarios: the safety-critical data, which represents the most challenging driving situation, and the general dataset, which covers a wider range of driving conditions. The comparative results shown below are intended to illustrate the impact of our dataset enhancement techniques on both the original dataset and other benchmark methods such as SMOGN and importance sampling.

\begin{table}[h]
\centering
\caption{Comparison of Model Performance on Original and Augmented Datasets}
\label{tab:results_vertical}
\begin{tabular}{|c|c|c|}
\hline
\multicolumn{1}{|c|}{\textbf{Method}} & \textbf{safety-critical data} & \textbf{Complete Dataset} \\ \cline{2-3}
\multicolumn{1}{|c|}{} & \textbf{RMSE} / \textbf{MAE} & \textbf{RMSE} / \textbf{MAE} \\ \hline
\textbf{Original Dataset} & 1.8725 / 1.6504 & 0.2416 / 0.1217 \\ \hline
\textbf{SMOGN} & 1.6966 / 1.4883 & 0.4256 / 0.2612 \\ \hline
\textbf{Importance Sampling} & 1.7934 / 1.5635 & 0.2107 / 0.1547 \\ \hline
\textbf{Ours} & 1.6923 / 1.4312 & 0.2039 / 0.1132 \\ \hline
\end{tabular}
\end{table}

\begin{figure}
    \centering
    \includegraphics[width=1\linewidth]{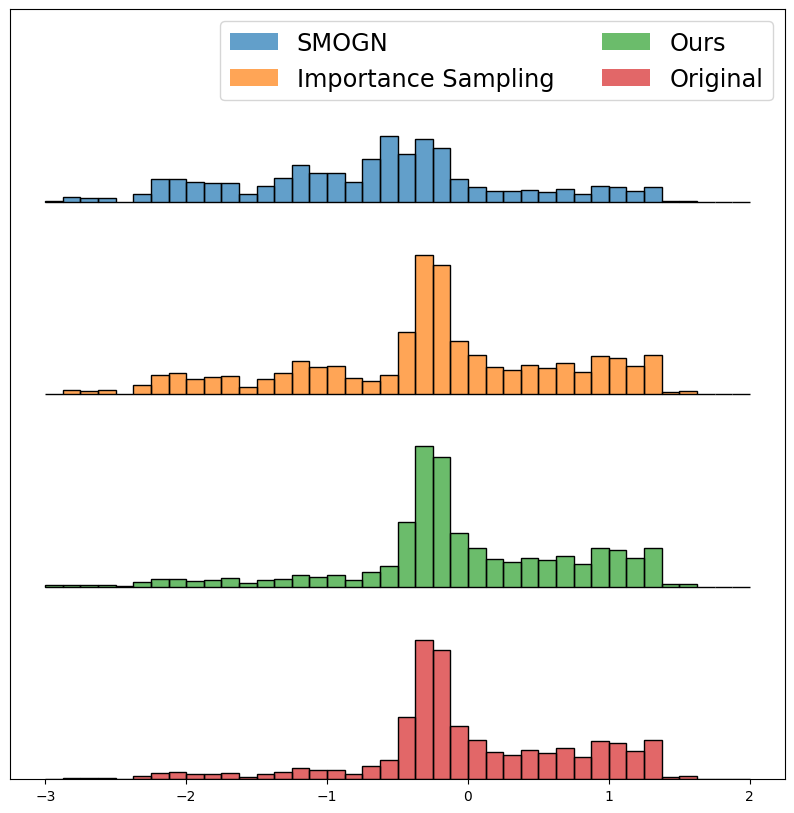}
    \caption{Distribution of Acceleration}
    \label{fig:dis}
\end{figure}

In Table \ref{tab:results_vertical}, our augmented dataset is shown to be effective in improving the accuracy of the model, especially when dealing with the safety-critical data. The observed reductions in both root mean square error (RMSE) and mean absolute error (MAE) under the safety-critical data indicate that the model trained with the augmented data is more capable of accurately predicting vehicle acceleration under hazardous conditions. This improvement is critical for autonomous driving systems, whose prediction accuracy has a direct impact on safety and operational efficiency. In addition, the performance metrics for the full dataset show that instead of decreasing the overall predictive ability of the model, the inclusion of augmented data enhances the overall predictive ability of the model under typical driving conditions. The slight improvement in RMSE and MAE for the general dataset validates the role of the augmented dataset in improving model robustness. Furthermore, Figure \ref{fig:dis} demonstrates the distribution of the dataset after processing by the various methods. This figure clearly shows that our augmentation had a minimal overall impact on the data, maintaining the authentic characteristics of the original dataset while enhancing it to better represent safety-critical scenarios.

\section{conclusion}

This study effectively demonstrates a new approach to enhance the driving dataset by simulating more dangerous driving scenarios, thereby addressing the problem of unbalance in safety critical data in the autonomous driving training dataset. By integrating vehicle detection, depth estimation, and 3D modeling techniques, we successfully created and modified images to increase the occurrence of these challenging situations. Training of the model on both the original and augmented datasets showed that the inclusion of these modified images significantly improved the model's performance in high-risk scenarios, while accuracy in normal conditions was not compromised.

\bibliographystyle{IEEEtran}
\bibliography{ieee.bib}

\end{document}